\documentclass[10pt,twocolumn,letterpaper]{article}

\usepackage{iccv}
\usepackage{times}
\usepackage{epsfig}
\usepackage{graphicx}
\usepackage{amsmath}
\usepackage{amssymb}
\usepackage[accsupp]{axessibility}

\usepackage{tabularx}
\usepackage{amssymb}
\usepackage{booktabs}
\usepackage{color}

\newcommand{\figVspaceUp}{-8pt} % vpsace for upper figure caption
\newcommand{\figVspaceDown}{-9pt}
\newcommand{\tabVspaceUp}{3pt} % vpsace for upper table caption
\newcommand{\tabVspaceDown}{-11pt}
\newcommand{\secVspace}{-2pt}
\newcommand{\subsecVspace}{-4pt}
\newcommand{\subsubsecVspaceUp}{-4pt}
\newcommand{\subsubsecVspace}{-4pt}
\usepackage[pagebackref=true,breaklinks=true,colorlinks,bookmarks=false]{hyperref}

\iccvfinalcopy % *** Uncomment this line for the final submission

\def\iccvPaperID{8106} % *** Enter the ICCV Paper ID here

\ificcvfinal\fi

\begin{document}

%%%%%%%%% TITLE
\title{GyroFlow: Gyroscope-Guided Unsupervised Optical Flow Learning}

\author{
    Haipeng Li$^1$ \quad \
    Kunming Luo$^1$ \quad \ 
    Shuaicheng Liu$^{2,1}$\thanks{Corresponding author}  \\
    \\
    $^1$Megvii Technology \\
    $^2$University of Electronic Science and Technology of China\\

{\tt\small \{lihaipeng, luokunming\}@megvii.com}, {\tt\small liushuaicheng@uestc.edu.com}

}

\maketitle
\ificcvfinal\thispagestyle{empty}\fi

%%%%%%%%% ABSTRACT

\begin{abstract}
Existing optical flow methods are erroneous in challenging scenes, such as fog, rain, and night because the basic optical flow assumptions such as brightness and gradient constancy are broken. To address this problem, we present an unsupervised learning approach that fuses gyroscope into optical flow learning. Specifically, we first convert gyroscope readings into motion fields named gyro field. Second, we design a self-guided fusion module to fuse the background motion extracted from the gyro field with the optical flow and guide the network to focus on motion details. To the best of our knowledge, this is the first deep learning-based framework that fuses gyroscope data and image content for optical flow learning. To validate our method, we propose a new dataset that covers regular and challenging scenes. Experiments show that our method outperforms the state-of-art methods in both regular and challenging scenes. Code and dataset are available at \href{https://github.com/megvii-research/GyroFlow}{https://github.com/megvii-research/GyroFlow}.
\end{abstract}

%%%%%%%%% BODY TEXT
\section{Introduction}
\vspace{\secVspace}

Optical flow estimation is a fundamental yet essential computer vision task that has been widely applied in various applications such as object tracking~\cite{behl2017bounding}, visual odometry~\cite{campbell2004techniques}, and image alignments~\cite{kroeger2016fast}. The original formulation of the optical flow was proposed by Horn and Schunck~\cite{horn1981determining}, after which the accuracy of optical flow estimation algorithms has been improved steadily. Early traditional methods minimize pre-defined energy functions with various assumptions and constraints~\cite{lucas1981iterative}. Deep learning-based methods directly learn the per-pixel regression through convolutional neural networks, which can be divided into supervised~\cite{dosovitskiy2015flownet, ranjan2017optical, teed2020raft} and unsupervised methods~\cite{ren2017unsupervised, luo2021upflow}. The former methods are primarily trained on synthetic data~\cite{dosovitskiy2015flownet, butler2012naturalistic} due to the lack of ground-truth labels. In contrast, the later ones can be trained on abundant and diverse unlabeled data by minimizing the photometric loss between two images. Although existing methods achieve good results, they rely on image contents, requiring images to contain rich texture and similar illumination conditions.

\begin{figure}[t]
\begin{center}
  \includegraphics[width=1\linewidth]{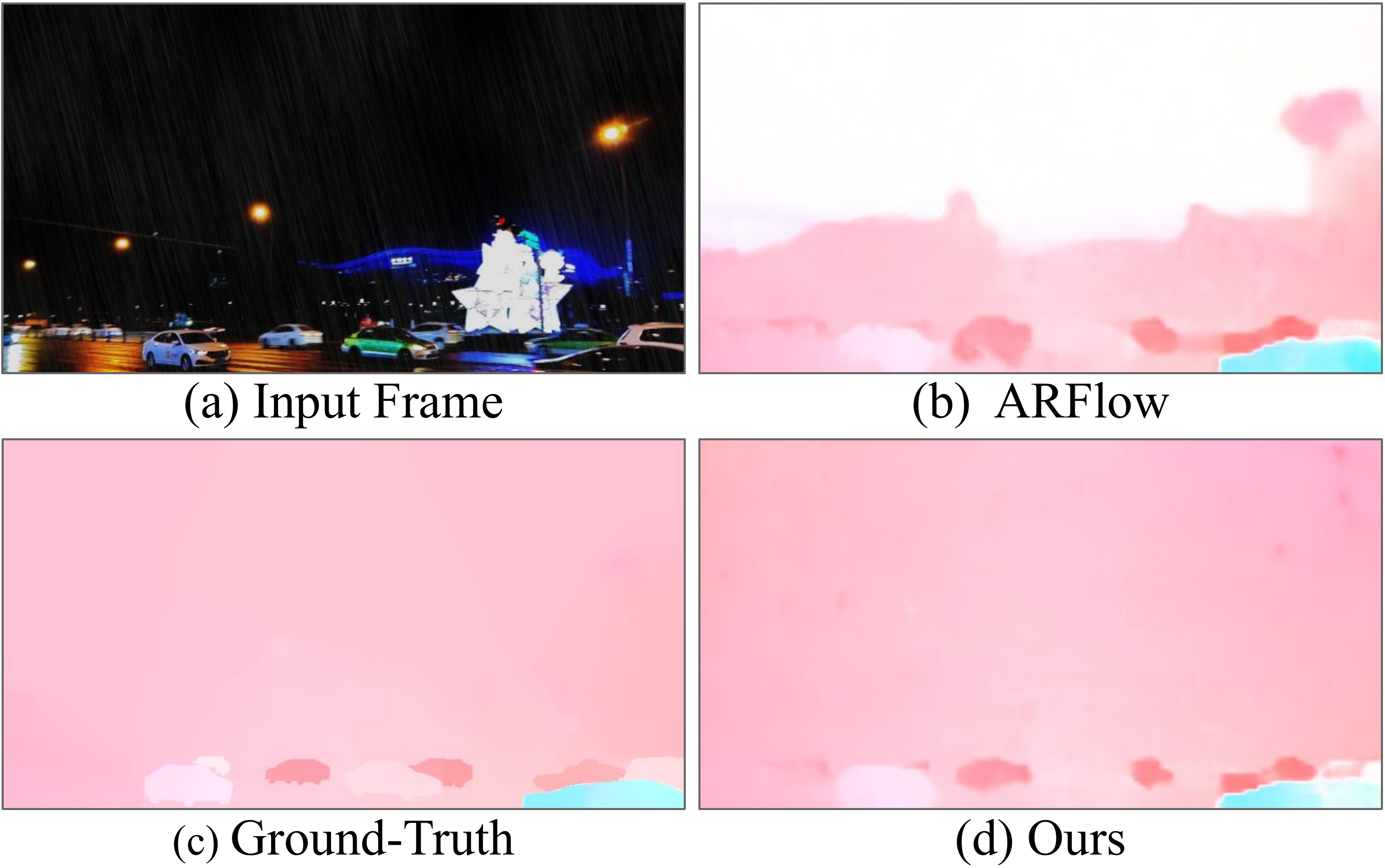}
\end{center}
  \vspace{\figVspaceUp}
  \caption{(a) Input low-light frame. (b) Optical flow result from existing baseline method ARFlow~\cite{liu2020learning}. (c) Ground-Truth. (d) Result from our GyroFlow.}
  \vspace{\figVspaceDown}
\label{fig:teaser}
\end{figure}

% On the other hand, a gyroscope can provide 3D motion in terms of roll, pitch and yaw, which has been widely used for the system control~\cite{} and the HCI of mobiles~\cite{}. Among all the potential possibilities~\cite{}({\textcolor{red}{Cite some gyro-based applications, at least 3}}), one is to fuse the gyroscope for the camera motion estimation. If a gyroscope works with a camera, it can measure the camera rotational motion. The gyro readings can be then converted into motion fields that describes the background motion given the camera intrinsic parameters. It is attractive that a gyroscope does not look at the image contents, but can produce reliable background motion. Not matter what the image contents are, e.g., either poor texture or dynamic scenes, the gyroscope can continuously provide reliable and accurate global camera motion. 

% However, the drawback is also obvious, the motion provided by a gyroscope is limited to the camera rotational motion. The translational motion cannot be recovered. Moreover, motion of dynamic objects are also absent. 

On the other hand, gyroscopes do not rely on image contents, which provide angular velocities in terms of roll, pitch, and yaw that can be converted into $3$D motion, widely used for system control~\cite{leland2006adaptive} and the HCI of mobiles~\cite{gupta2016continuous}. Among all potential possibilities~\cite{bloesch2014fusion, li2018efficient, hwangbo2009inertial}, one is to fuse the gyro data for the motion estimation. Hwangbo~\emph{et al.} proposed to fuse gyroscope to improve the robustness of KLT feature tracking~\cite{hwangbo2009inertial}. Bloesch~\emph{et al.} fused gyroscope for the ego-motion estimation~\cite{bloesch2014fusion}. These attempts demonstrate that if the gyroscope is integrated correctly, the performance and the robustness of the method can be largely improved.   

Given camera intrinsic parameters, gyro readings can be converted into motion fields to describe background motion instead of dynamic object motion because it is confined to camera motion. It is engaging that gyroscopes do not require the image contents but still produce reliable background camera motion under conditions of poor texture or dynamic scenes. Therefore, gyroscopes can be used to improve the performance of optical flow estimation in challenging scenes, such as poor texture or inconsistent illumination conditions.

% \textcolor{red}{todo draw a dragon eye}

% However, the drawback is also obvious, motion recorded by the gyroscope are limited to the camera rotations. Camera translational motion cannot be recovered. Meanwhile, motion from dynamic objects and scene depth are also absent.

In this paper, we propose GyroFlow, a gyroscope-guided unsupervised optical flow estimation method. We combine the advantages of image-based optical flow that recovers motion details based on the image content with those of a gyroscope that provides reliable background camera motion independent of image contents. Specifically, we first convert gyroscope readings into gyro fields that describe background motion given the image coordinates and the camera intrinsic. Second, we estimate optical flow with an unsupervised learning framework and insert a proposed \textbf{S}elf-\textbf{G}uided \textbf{F}usion (SGF) module that supports the fusion of the gyro field during the image-based flow calculation. Fig.~\ref{fig:teaser} shows an example, where Fig.~\ref{fig:teaser} (a) represents the input of a night scene with poor image texture, and Fig.~\ref{fig:teaser} (c) is the ground-truth optical flow between two frames. Image-based methods such as ARFlow~\cite{liu2020learning} (Fig.~\ref{fig:teaser} (b)) can produce the dynamic object motion but fail to compute the background motion in the sky, where no texture is available. Fig.~\ref{fig:teaser} (d) shows our GyroFlow fusion result. As seen, both global motion and motion details can be retained. From experiments, we notice that motion details can be better recovered if global motion is provided. 

% The fusion of two motion sources are non-trivial. Limited improvements or even drawbacks could be introduced if fused inappropriately.

% Specifically, we first convert gyroscope readings into gyro fields that describe background motion given the image coordinates and the camera intrinsic. Second, we estimate optical flow with an unsupervised learning framework, and insert a proposed \textbf{S}elf-\textbf{G}uided \textbf{F}usion (SGF) module that supports the fusion of gyro field during the image-based flow calculation. 

To validate our method, we propose a dataset GOF (\textbf{G}yroscope \textbf{O}ptical \textbf{F}low) containing scenes under 4 different categories with synchronized gyro readings, including one regular scene (RE) and three challenging cases as low light scenes (Dark), foggy scenes (Fog), and rainy scenes (Rain). For quantitative evaluations, we further propose a test set, which includes accurate optical flow labels by the method~\cite{liu2008human}, through extensive efforts. Note that existing flow datasets, such as Sintel~\cite{butler2012naturalistic}, KITTI~\cite{geiger2012we,menze2015object} cannot be used for the evaluation due to the absence of the gyroscope readings. To sum up, our main contributions are:

\begin{itemize}
    \item We propose the first DNN-based framework that fuses gyroscope data into optical flow learning.
    \item We propose a self-guided fusion module to effectively realize the fusion of gyroscope and optical flow.
    \item We propose a dataset for the evaluation. Experiments show that our method outperforms existing methods.
\end{itemize}

% As illustrated in Fig.~\ref{fig:gyro_pipe}, given start timestamps $t_{I_{a}}$ and $t_{I_{b}}$ of frames $I_{a}$ and $I_{b}$, the relative gyroscope readings can be converted into gyro fields $G_{ab}$. Then we feed $I_{a}$, $I_{b}$ and $G_{ab}$ to our GyroFlow to produce an output flow $V_{ab}$ as shown in Fig.~\ref{fig:pipeline}. 

\section{Related Work\label{sec:related}}
\vspace{\secVspace}
\subsection{Gyro-based Vision Applications}
\vspace{\subsecVspace}
Gyroscopes reflect the camera rotation. Many applications equipped with the gyroscope have been widely applied, including but not limited to video stabilization~\cite{karpenko2011digital}, image deblur~\cite{mustaniemi2019gyroscope}, optical image stabilizer (OIS)~\cite{la2015optical}, simultaneous localization and mapping (SLAM)~\cite{huang2018online}, ego-motion estimation~\cite{bloesch2014fusion}, gesture-based user authentication on mobile devices~\cite{guse2012gesture}, image alignment with OIS calibration~\cite{li2021deepois} and human gait recognition~\cite{zhang2004human}. The gyroscopes are important in mobile phones. The synchronization between the gyro readings and the video frames is important. Jia~\emph{et al.}~\cite{jia2013online} proposed gyroscope calibration to improve the synchronization. Bloesch~\emph{et al.}~\cite{bloesch2014fusion} fused optical flow and inertial measurements to deal with the drifting issue. In this work, we acquire gyroscope data from the bottom layer of the Android layout, i.e., Hardware Abstraction Layer (HAL), to achieve accurate synchronizations.

\subsection{Optical Flow}
\vspace{\subsecVspace}
Our method is related to optical flow estimation. Traditional methods minimize the energy function between image pairs to compute an optical flow~\cite{lucas1981iterative}. Recent deep approaches can be divided into supervised~\cite{dosovitskiy2015flownet, ranjan2017optical, teed2020raft} and unsupervised methods~\cite{ren2017unsupervised, luo2021upflow}. 

Supervised methods require labeled ground-truth to train the network. FlowNet~\cite{dosovitskiy2015flownet} first proposed to train a fully convolutional network on synthetic dataset FlyingChairs. To deal with the large displacement scenes, SpyNet~\cite{ranjan2017optical} introduced a coarse-to-fine pyramid network. PWC-Net~\cite{sun2018pwc}, LiteFlowNet~\cite{hui2018liteflownet}, IRR-PWC~\cite{hur2019iterative} designed lightweight and efficient networks by warping features, computing cost volumes, and introducing residual learning for iterative refinement with shared weights. Recently, RAFT~\cite{teed2020raft} achieved state-of-the-art performance by constructing a pixel-level correlation volume and using a recurrent network to estimate optical flow. 

Unsupervised methods do not require ground-truth annotations. DSTFlow~\cite{ren2017unsupervised} and Back2Basic~\cite{jason2016back} are pioneers for unsupervised optical flow estimation. Several works~\cite{meister2018unflow, liu2019ddflow, wang2018occlusion, liu2020learning} focus on dealing with the occlusion problem by forward-backward occlusion checking, range-map occlusion checking, data distillation, and augmentation regularization loss. Other methods concentrate on optical flow learning by improving image alignment, including the census loss~\cite{meister2018unflow}, formulation of multi-frames~\cite{janai2018unsupervised}, epipolar constraints~\cite{zhong2019unsupervised}, depth constraints~\cite{yin2018geonet}, feature similarity constraints~\cite{im2020unsupervised}, and occlusion inpainting~\cite{liu2021oiflow}. UFlow~\cite{jonschkowski2020matters} proposed a unified framework to systematically analyze and integrate different unsupervised components. Recently, UPFlow~\cite{luo2021upflow} proposed a neural upsampling module and pyramid distillation loss to improve the upsampling and learning of pyramid network, achieving state-of-art performance. 

However above methods may not work well under challenging scenes, such as dark, rain, and fog environments. Zheng~\emph{et al.} proposed a data-driven method that establishes a noise model to learn optical flow from low-light images~\cite{zheng2020optical}. Li~\emph{et al.} proposed a RainFlow, which includes $2$ modules to handle the rain veiling effect and rain streak effect respectively, to produce optical flow in the heavy rain~\cite{li2019rainflow}. Yan~\emph{et al.} proposed a semi-supervised network that converts foggy images into clean images to deal with dense foggy scenes~\cite{yan2020optical}. In this paper, we build our GyroFlow upon unsupervised components with the fusion of gyroscope to cover both regular and the challenging scenes.

\begin{figure*}[t]
\begin{center}
  \includegraphics[width=0.95\linewidth]{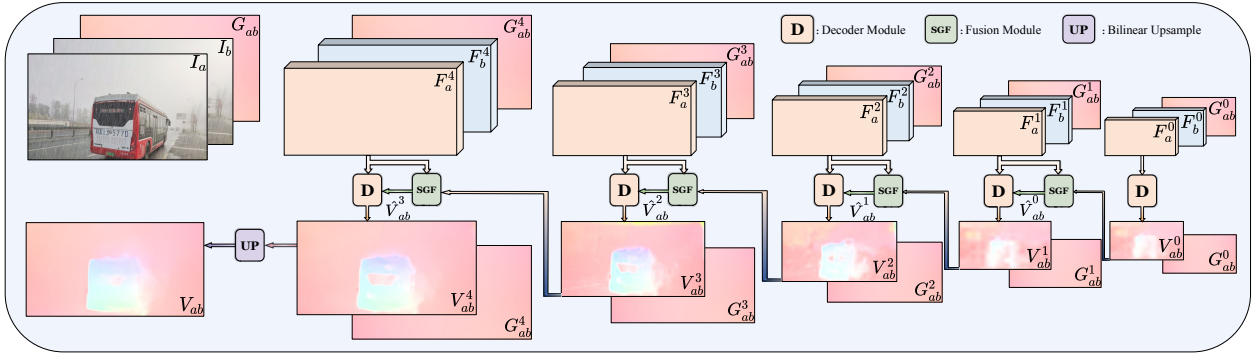}
\end{center}
%   \caption{The overview of our algorithm. It makes up of a pyramid encoder and a pyramid decoder. For each pair of frames $I_{a}$ to $I_{b}$, the encoder extracts features at different scales, and the decoder includes two modules, for the pyramid layer $l$,  of which the first functions to fuse a gyro field $G_{ab}^{l}$ and an optical flow $V_{ab}^{l}$ to produce a fused flow as input to the second module which estimates an optical flow.}
\vspace{\figVspaceUp}
  \caption{The overview of our algorithm. It consists of a pyramid encoder and a pyramid decoder. For each pair of frames $I_{a}$ to $I_{b}$, our encoder extracts features at different scales. The decoder includes two modules, at each layer $l$, $\mathbf{SGF}$ functions to fuse a gyro field $G_{ab}^{l}$ and an optical flow $V_{ab}^{l}$ to produce a fused flow $\hat{V_{ab}^{l}}$ as input to $\mathbf{D}$, which estimates an optical flow to the next layer.}
\label{fig:pipeline}
\vspace{\figVspaceDown}
\end{figure*}

\subsection{Gyro-based Motion Estimation}
\vspace{\subsecVspace}
% {\color{red}{Describe some traditional works that fuse the gyro and flow. You mentioned that there do exist some works that fuse the gyro data for motion estimation, although not directly for the optical flow. But we should describe them explicitly here in the form of, 

% e.g., Zhangsan~\emph{et al.} prospoed,  Lisi~\emph{et al.} adopted.., However, none of them fuse the gyro data for the optical flow improvements. In this work, we propose a DNN-based solution that fuse the gyro data with image-based optical flow. blar blar blar..... }}

Hwangbo~\emph{et al.} proposed an inertial-aided KLT feature tracking method to handle the camera rolling and illumination change~\cite{hwangbo2009inertial}. Bloesch~\emph{et al.} presented a method for fusing optical flow and inertial measurements for robust ego-motion estimation~\cite{bloesch2014fusion}. Li~\emph{et al.} proposed a gyro-aided optical flow estimation method to improve the performance under fast rotations~\cite{li2018efficient}. Specifically, they produce a sparse optical flow that ignores foreground motion. However, none of them took challenging scenes into account nor used neural networks to fuse gyroscope data for optical flow improvement. In this work, including producing dense optical flow and taking rolling-shutter effects into account, we propose a DNN-based solution that fuses gyroscope data to image-based flow to improve optical flow estimations.
% {\textcolor{red}{Why we cannot compare with these traditional methods, implementation issue ? not exactly the same ? or they designed for a different purpose?  Anyway, we need some explanations. Maybe here, but more importantly, in the experiment section.  }}

\section{Algorithm}
\vspace{\secVspace}
Our method is built upon convolutional neural networks that inputs a gyro field $G_{ab}$ and two frames $I_{a}$, $I_{b}$ to estimate a forward optical flow $V_{ab}$ that describes the motion for every pixel from $I_{a}$ towards $I_{b}$ as:
\begin{equation}
\small
% V_{ab}=\mathcal{F}\left(\theta, G_{ab}, I_{a}, I_{b}\right),
V_{ab} = \mathcal{F}_{\theta}\left(G_{ab}, I_a, I_b\right),
\end{equation}
where $\mathcal{F}$ is our network with parameter $\theta$.

% Our pipeline consists of two modules: a gyro field estimator and a network to fuse images and the gyro field.
Fig.~\ref{fig:pipeline} illustrates our pipeline. Firstly, the gyro field $G_{ab}$ is produced by the gyroscope readings between the relative frames $I_{a}$ and $I_{b}$ (Sec.~\ref{sec:gyro_field}), then it is concatenated with the two frames to be fed into the network to produce an optical flow $V_{ab}$ between $I_{a}$ and $I_{b}$. Our network consists of two stages. For the first stage, we extract feature pairs at different scales. For the second stage, we use the decoder $\mathbf{D}$ and the self-guided fusion module $\mathbf{SGF}$ (Sec.~\ref{sec:SFG}) to produce optical flow in a coarse-to-fine manner.

Our decoder $\mathbf{D}$ is same as UPFlow~\cite{luo2021upflow} which consists of the feature warping~\cite{sun2018pwc}, the cost volume construction~\cite{sun2018pwc}, the cost volume normalization~\cite{jonschkowski2020matters}, the self-guided upsampling~\cite{luo2021upflow}, and the parameter sharing~\cite{hur2019iterative}. In summary, the second pyramid decoding stage can be formulated as:

\begin{equation}
\small
\begin{aligned}
\hat{V}_{ab}^{i-1} &=\mathbf{SGF}\left(F_{a}^{i}, F_{b}^{i},V_{ab}^{i-1},G_{ab}^{i-1}\right), \\
V_{ab}^{i} &=\mathbf{D}\left(F_{a}^{i}, F_{b}^{i}, \hat{V}_{ab}^{i-1}\right),
\end{aligned}
\end{equation}
where $i$ represents the number of pyramid levels, $F_{a}^{i}$, $F_{b}^{i}$ are extracted features from $I_{a}$ and $I_{b}$ at the $i$-th pyramid level. In the $i$-th layer, $\mathbf{SGF}$ takes image features $F_{a}^{i}$, $F_{b}^{i}$ from the feature pyramid, the output $V_{ab}^{i-1}$ of decoder $\mathbf{D}$ from the last layer and the downscale gyro field $G_{ab}^{i-1}$ as inputs, then it produces a fusion result $\hat{V}_{ab}^{i-1}$ which is fed into $\mathbf{D}$. The D takes image features $F_{a}^{i}$, $F_{b}^{i}$ and the fusion result $\hat{V}_{ab}^{i-1}$ as inputs and outputs a flow $V_{ab}^{i}$. Specifically, the output flow is directly upsampled at the last layer. Next, we first describe how to convert the gyro readings into a gyro field in Sec.~\ref{sec:gyro_field} and then introduce our $\mathbf{SGF}$ module in Sec.~\ref{sec:SFG}.

\subsection{Gyro Field}\label{sec:gyro_field}
\vspace{\subsecVspace}

\begin{figure}[t]
\begin{center}
  \includegraphics[width=1\linewidth]{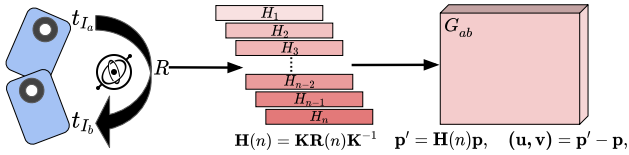}
\end{center}
\vspace{\figVspaceUp}
%   \caption{The pipeline of generating gyro field. For each frame pair, we record the timestamp of the first frame $t_{I_a}$ and the second frame $t_{I_{b}}$. Then the gyroscope readings can be read out to compute an array of rotation matrix $R=\left(R_{1} \ldots R_{n}\right)^\mathsf{T}$ in the case of rolling-shutter camera. Furthermore, we convert the rotation array into the homography array that projects pixels $p$ of the first image into $p^{\prime}$, yielding a gyro field $G_{ab}$ by subtracting $p^{\prime}$ and $p$.}
  \caption{The pipeline of generating gyro field. Given timestamps $t_{I_a}$ and $t_{I_{b}}$, gyroscope readings can be read out to compute an array of rotation matrices $R=\left(R_{1} \ldots R_{n}\right)^\mathsf{T}$. We then convert the rotation array into the homography array that projects pixels $p$ of the first image into $p^{\prime}$, yielding a gyro field $G_{ab}$.}
\vspace{\figVspaceDown}
\label{fig:gyro_pipe}
\end{figure}

We obtain gyroscope readings from mobile phones that are widely available and easy to access. For mobile phones, gyroscopes reflect camera rotations. We compute rotations by compounding gyroscope readings that include 3-axis angular velocities and timestamps. In particular, compared to previous work~\cite{karpenko2011digital, kundra2014bias, mustaniemi2019gyroscope} that read gyro readings from the API, we directly read them from HAL of Android architecture to avoid the non-trivial synchronization problem that is critical for the gyro accuracy. Between frames $I_a$ and $I_b$, the rotation vector $n=\left(\omega_{x}, \omega_{y}, \omega_{z}\right) \in \mathbb{R}^{3}$ is computed according to method~\cite{karpenko2011digital}, then the rotation matrix $R(t)\in SO(3)$ can be produced by Rodrigues Formula~\cite{dai2015euler}.

In the case of a global shutter camera, e.g., the pinhole camera, a rotation-only homography can be computed as:
\begin{equation}
\small
\mathbf{H(t)}=\mathbf{K} \mathbf{R}(t) \mathbf{K}^{-1}\label{eq:globalH},
\end{equation}
\noindent where $K$ is the camera intrinsic matrix, $t$ represents the time from the first frame $I_{a}$ to the second frame $I_{b}$, and $R(t)$ denotes the camera rotation from $I_a$ to $I_b$.

For a rolling shutter camera that most mobile phones adopt, each scanline of the image is exposed at a slightly different time, as illustrated in Fig~\ref{fig:gyro_pipe}. Therefore, Eq.~(\ref{eq:globalH}) is not applicable anymore, since every row of the image should have a different orientation. In practice, it is not necessary to assign each row with a rotation matrix. We group several consecutive rows into a row patch and assign each patch with a rotation matrix. The number of row patches depends on the number of gyroscope readings per frame.

% \begin{equation}
% \small
% t_{a}(i)=t_{I}+t_{s} \frac{i}{N},
% \end{equation}
% \noindent where $t_{a}(i)$ denotes the start of the exposure of the $i$-$th$ patch in $I_a$ as shown in Fig.~\ref{fig:gyro_ts}, $t_{I}$ denotes the starting timestamp of the corresponding frame, $N$ denotes the number of patches per frame. The end of the exposure is:

% \begin{equation}
% \small
% t_{b}(i)=t_{a}(i)+t_{f},
% \end{equation}
% \noindent where $t_{f}=1/FPS$ is the frame period. 
Here, the homography between the $n$-th row at frame $I_a$ and $I_b$ can be modeled as:

\begin{equation}
\small
\mathbf{H}_{n}(t)=\mathbf{K} \mathbf{R}\left(t_{b}\right) \mathbf{R}^{\top}\left(t_{a}\right) \mathbf{K}^{-1},
\end{equation}
where the $n$ is the index of row patches, $\mathbf{H}_{n}(t)$ denotes the homography of the $n$-th row patch from $I_{a}$ to $I_{b}$, and $\small\mathbf{R}\left(t_{b}\right) \mathbf{R}^{\top}\left(t_{a}\right)$ can be computed by accumulating rotation matrices from $t_a$ to $t_b$.

In our implementation, we regroup the image into $14$ patches that compute a homography array containing $14$ horizontal homography between two consecutive frames. Furthermore, to avoid the discontinuities across row patches, we convert the array of homography into an array of $4$D quaternions~\cite{zhang1997quaternions} and then apply the spherical linear interpolation (SLERP) to interpolate the camera orientation smoothly, yielding a smooth homography array. As shown in Fig~\ref{fig:gyro_pipe}, we use the homography array to transform every pixel $p$ to $p^{\prime}$, and subtract $p^{\prime}$ from $p$ as:

\begin{equation}
\small
\mathbf{p}^{\prime}=\mathbf{H(n)} \mathbf{p}, \quad \mathbf{(u,v)}=\mathbf{p}^{\prime}- \mathbf{p},
\label{eq:homo2flow}
\end{equation}
\noindent computing offsets for every pixel produces a gyro field $G_{ab}$.

\subsection{Self-guided Fusion Module}\label{sec:SFG}
\vspace{\subsecVspace}
As Fig.~\ref{fig:teaser} illustrates: Fig.~\ref{fig:teaser} (a) denotes the input images. Fig.~\ref{fig:teaser} (b) is the output of the ARFlow~\cite{liu2020learning}, an unsupervised optical flow approach, where only the motion of moving objects is roughly produced. As image-based optical flow methods count on image contents for the registration, they are prone to be erroneous in challenging scenes, such as textureless scenarios, dense foggy environments~\cite{yan2020optical}, dark~\cite{zheng2020optical} and rainy scenes~\cite{li2019rainflow}. Fig.~\ref{fig:teaser} (c) represents the ground-truth. To combine the advantages of the gyro field and the image-based optical flow, we propose a self-guided fusion module (SGF). In Fig.~\ref{fig:teaser} (d), with the gyro field, our result is much better compared with the ARFlow~\cite{liu2020learning}. 

\begin{figure}[t]
\begin{center}
  \includegraphics[width=1\linewidth]{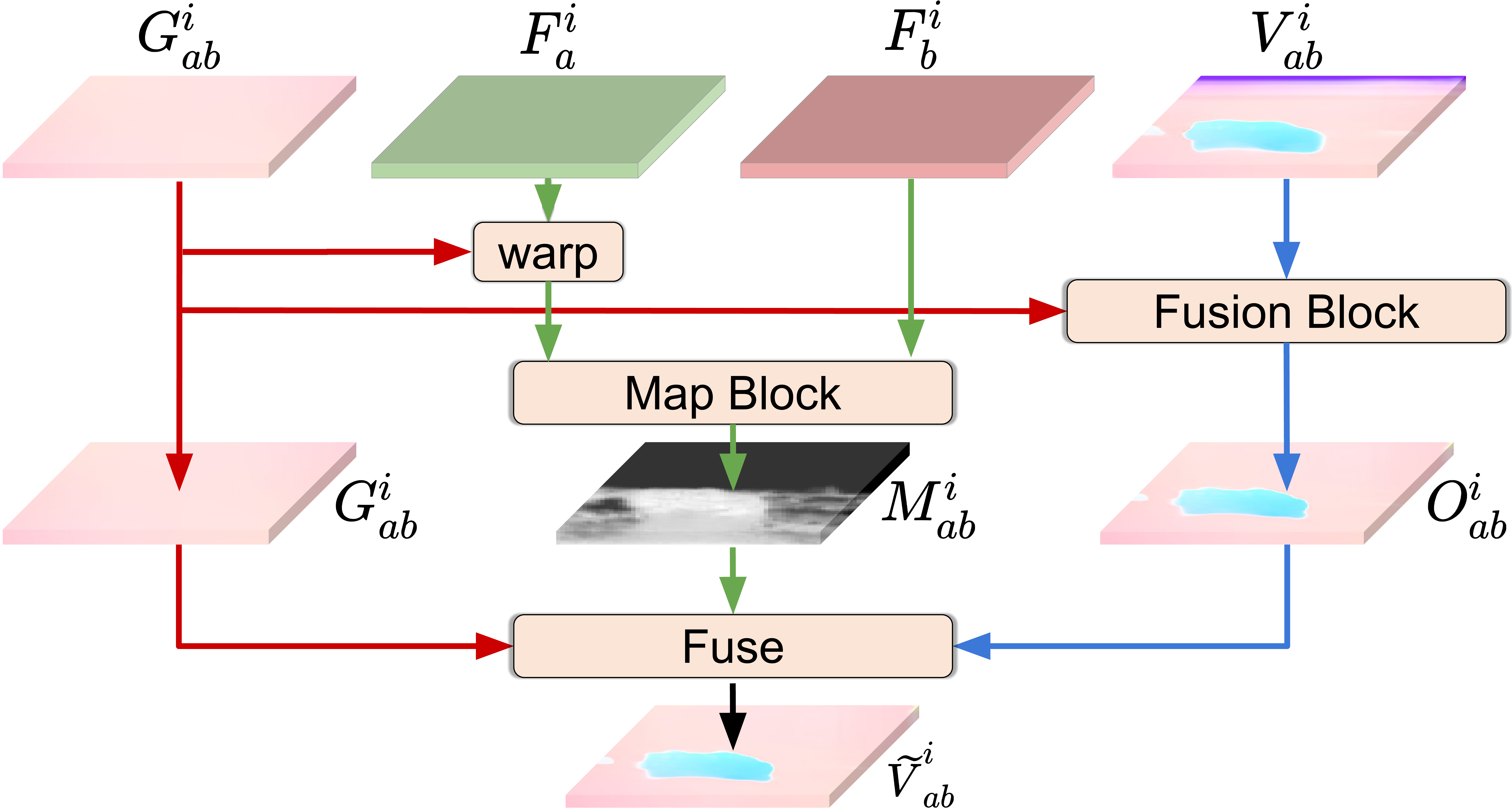}
\end{center}
\vspace{\figVspaceUp}
  \caption{Illustration of our self-guided fusion module (SGF). For a specific layer $i$, we use $2$ blocks to independently produce the fusion map $M^{i}_{ab}$ and the fusion flow $O^{i}_{ab}$, then we generate the output $\widetilde{V}_{ab}^{i}$ by Eq.~\ref{fuse_eq}.}
\label{fig:SGF}
\vspace{\figVspaceDown}
\end{figure}

The architecture of our SGF is shown in Fig.~\ref{fig:SGF}. Given the input features of image $I_a$ and $I_b$ at the $i$-th layer as $F^{i}_{a}$ and $F^{i}_{b}$. $F^{i}_{a}$ is warped by the gyro field $G^{i}_{ab}$, which is the forward flow from feature $F^{i}_{a}$ to $F^{i}_{b}$. Then the warped feature is concatenated with $F^{i}_{b}$ as inputs to the map block, yielding a fusion map $M^{i}_{ab}$ that ranges from $0$ to $1$. Note that, in $M^{i}_{ab}$, those background regions which can be aligned with the gyro field are close to zeros, while the rest areas are distributed with different weights. Next, we input the gyro field $G^{i}_{ab}$ and optical flow $V^{i}_{ab}$ to the fusion block that computes a fusion flow $O^{i}_{ab}$. Finally, we fuse the $G^{i}_{ab}$ and $O^{i}_{ab}$ with $M^{i}_{ab}$ to guide the network to focus on the moving foreground regions. The process can be described as:

\begin{equation}
\small
    \widetilde{V}_{ab}^{i}=M_{ab}^{i} \odot O_{ab}^{i}+\left(1-M_{ab}^{i}\right) \odot G_{ab}^{i},
    \label{fuse_eq}
\end{equation}
where $\widetilde{V}_{ab}^{i}$ is the output of our SGF module and $\odot$ denotes the element-wise multiplier.

\section{Experimental Results}
\vspace{\secVspace}
\subsection{Dataset}
\vspace{\subsecVspace}
The representative datasets for optical flow estimation and evaluation include FlyingChairs~\cite{dosovitskiy2015flownet}, MPI-Sintel~\cite{butler2012naturalistic}, KITTI 2012~\cite{geiger2012we}, and KITTI 2015~\cite{menze2015object}. On the gyro side, a dedicated dataset embedded with gyroscopes, named GF4~\cite{li2021deepois}, is proposed for the homography estimation. However, none of them combine accurate gyroscope readings with image contents to evaluate the optical flow. Therefore, we propose a new dataset and benchmark: GOF.

\noindent\textbf{Training Set} Similar to GF4~\cite{li2021deepois}, a set of videos with gyroscope readings are recorded using a cellphone. Compared to GF4, which uses a phone with an OIS camera. We carefully choose a non-OIS camera phone to eliminate the effect of the OIS module. We collect videos in $4$ different environments, including regular scenes (RE), low light scenes (Dark), foggy scenes (Fog), and rainy scenes (Rain). For each scenario, we record a set of videos lasting for $60$ seconds, yielding $1,800$ frames under every environment. In total, we collect $5,000$ frames for the training set.

\begin{figure}[t]
\begin{center}
  \includegraphics[width=1\linewidth]{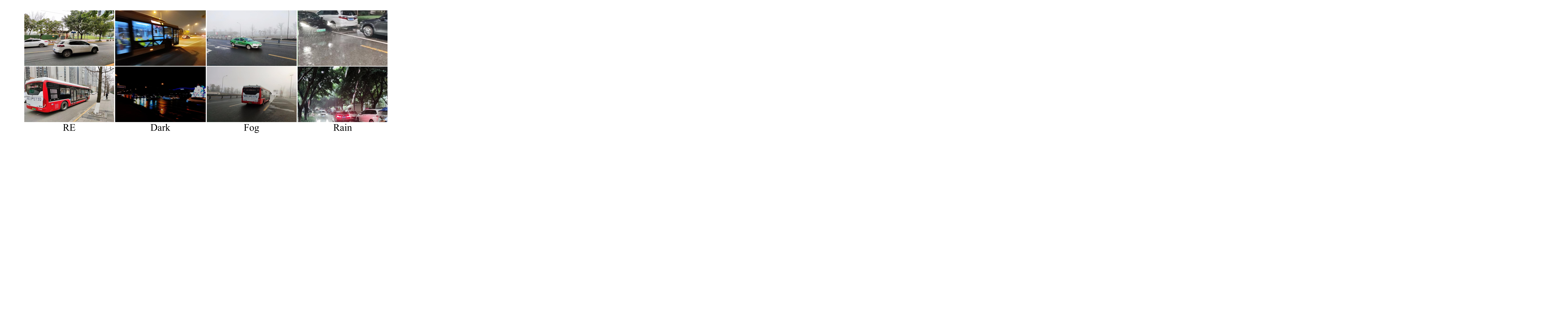}
\end{center}
\vspace{\figVspaceUp}
  \caption{A glance at our evaluation set. It can be divided into $4$ categories, including regular scenes(RE), low light scenes(Dark), foggy scenes(Fog), and rainy scenes(Rain). Each category contains $70$ pairs, and a total of $280$ pairs evaluation dataset is proposed with synchronized gyroscope readings.}
\label{fig:dataset}
\vspace{\figVspaceDown}
\end{figure}

\noindent\textbf{Evaluation Set} For evaluation, similar to the train set, we capture videos in $4$ scenes to compare to image-based registration methods. Each category contains $70$ pairs, yielding a $280$ pairs evaluation set. Fig.~\ref{fig:dataset} shows some examples.

% \vspace{-3mm}
% \vspace{-3mm}

% For quantitative evaluation, a ground-truth optical flow is required for each pair. However, labeling flow ground-truth is non-trivial. As far as we know, no powerful tool is available for this task. We adopt the most related approach~\cite{liu2008human} to this task, to label the ground-truth flow with many efforts, approximately $20~\sim 30$ minutes per image, especially for challenging scenes. For challenging scenes, we firstly label amout of $500$ examples, then we choose those with good visual performance and discard the others. Furthermore, we refine the selected samples with detail modification around the motion boundaries.
% including many switches between the flow and the image, and detailed modifications on the motion boundaries.

For quantitative evaluation, a ground-truth optical flow is required for each pair. However, labeling ground-truth flow is non-trivial. As far as we know, no powerful tool is available for this task. Following~\cite{li2019rainflow, yan2020optical}, we adopt the most related approach~\cite{liu2008human} to label the ground-truth flow with many efforts. It costs approximately $20\sim30$ minutes per image, especially for challenging scenes. We firstly label an amount of $500$ examples containing rigid objects, then we select those with good visual performance, i.e., the performance of image alignment, and discard the others. Furthermore, we refine the selected samples with detailed modifications around the motion boundaries.

\begin{figure}[t]
\begin{center}
  \includegraphics[width=1\linewidth]{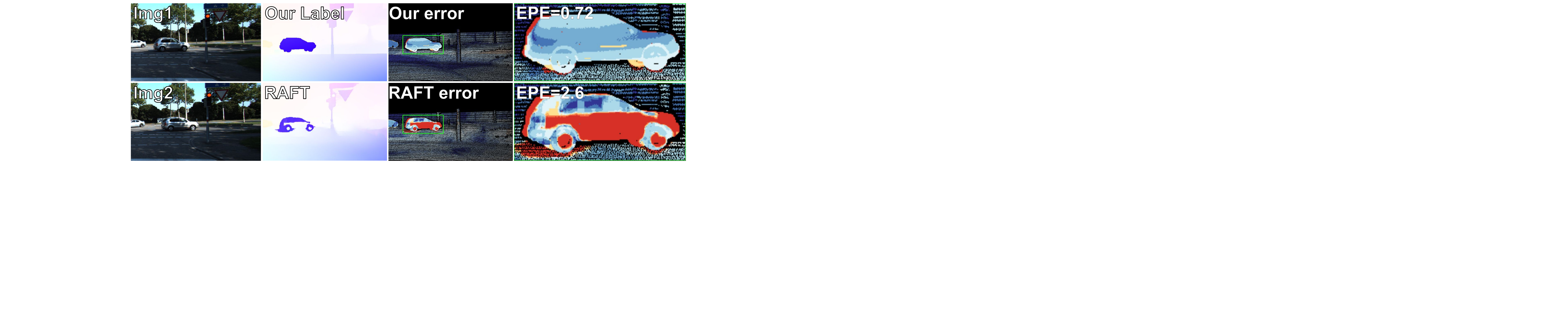}
\end{center}
\vspace{\figVspaceUp}
  \caption{One label example on KITTI 2012~\cite{geiger2012we}, compared to RAFT\cite{teed2020raft}(the second line) that computes an EPE equals $2.6$, our label flow(the first line) produces a $0.72$ EPE. From the error map, we notice that our labeled optical flow is much more accurate.}
\label{fig:liuce_gt}
% \vspace{-6pt}
\end{figure}

% To verify the effectiveness of our labeld optical flow, we choose to label several samples from KITTI 2012~\cite{geiger2012we}. Given the groundtruth, we compare our labeled optical flows with results produced by the state-of-the-art supervised method, i.e. RAFT~\cite{teed2020raft}. Our labeled flow computes an end point error (EPE) of $0.7$ where RAFT~\cite{teed2020raft} computes a EPE of $2.4$ which is more than $3$ times larger than ours. Fig.~\ref{fig:liuce_gt} shows the example. From the error map, we notice that our manually labeled flow is much more accurate than the current SOTA method. We use it as the groundtruth for evaluations. 
To verify the effectiveness of our labeled optical flow, we choose to label several samples from KITTI 2012~\cite{geiger2012we}. Given the ground-truth, we compare our labeled optical flow with results produced by the state-of-the-art supervised method, i.e., RAFT~\cite{teed2020raft} pre-trained on FlyingChairs. Our labeled flow computes an endpoint error (EPE) of $0.7$, where RAFT computes an EPE of $2.4$, which is more than $3$ times larger than ours. Fig.~\ref{fig:liuce_gt} shows one example. As the error map illustrates,  our labeled flow is much more accurate than the current SOTA method. We leverage this approach to generate ground-truth for evaluations.

\subsection{Implementation Details}
\vspace{\subsecVspace}
We conduct experiments on GOF dataset. Our method is built upon the PWC-Net~\cite{sun2018pwc}. For the first stage, we train our model for $100$k steps without the occlusion mask. For the second stage, we enable the bidirectional occlusion mask~\cite{meister2018unflow}, the census loss~\cite{meister2018unflow}, and the spatial transform~\cite{liu2020learning} to fine-tune the model for about $300$k steps.

We collect videos with gyroscope readings using Qualcomm QRD equipped with Snapdragon 7150, which records videos in $600 \times 800$ resolution. We add random crop, random horizontal flip, and random weather modification (add fog and rain~\cite{imgaug}) during the training. We report endpoint error (EPE) in the evaluation set. The implementation is in PyTorch, and one NVIDIA RTX 2080 Ti is used to train our network. We use Adam optimizer~\cite{kingma2014adam} with parameters setting as $LR=1.0 \times 10^{-4}$, $\beta_{1}=0.9$, $\beta_{2}=0.999$, $\varepsilon=1.0 \times 10^{-7}$. The batch size is $4$. It takes $3$ days to finish the entire training process. On one single $1080$ti, the time to generate an optical flow is $58$ms per frame. Same to previous work~\cite{jonschkowski2020matters, luo2021upflow}, we use the photometric loss and smooth term to train the network.

\begin{figure*}[t]
\begin{center}
  \includegraphics[width=1\linewidth]{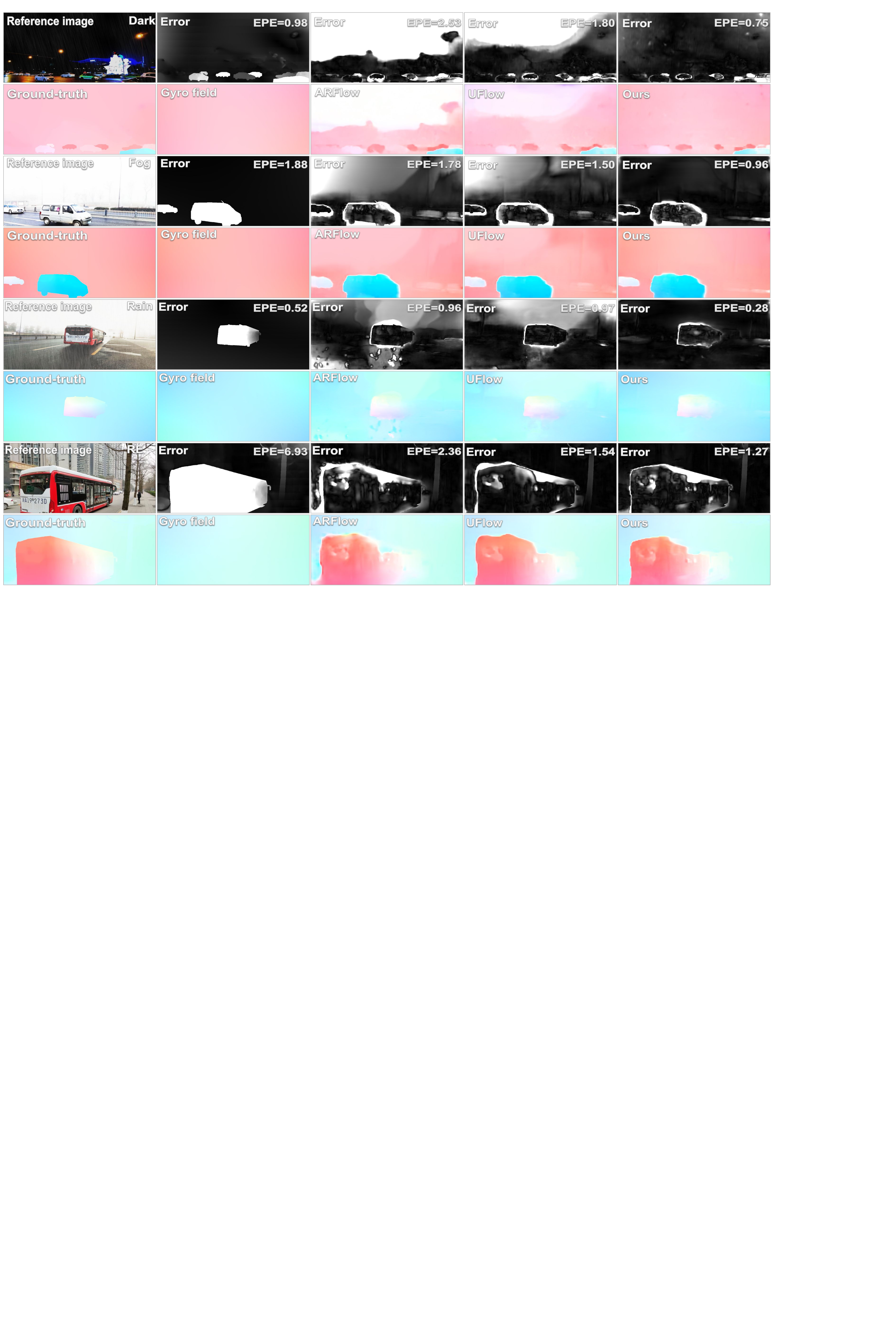}
\end{center}
\vspace{\figVspaceUp}
  \caption{Visual comparison of our method with gyro field, ARFlow~\cite{liu2020learning}, and UFlow~\cite{jonschkowski2020matters} on the GOF evaluation set. For the first $3$ challenging cases, we notice that our method achieves convincing results by fusing the background motion from the gyro field and the motion details from the optical flow. For the last example, in regular scenarios, fusing gyro field helps the learning of optical flow where the network produces accurate and sharp flow around the boundary of objects.}
\label{fig:qualitative}
\vspace{\figVspaceDown}
\end{figure*}

\subsection{Comparisons with Image-based Methods}
\vspace{\subsecVspace}
In this section, we compare our method with traditional, supervised, and unsupervised methods on GOF evaluation set with quantitative (Sec.~\ref{sec:quantitative}) and qualitative comparisons (Sec.~\ref{sec:qualitative}). To validate the effectiveness of key components, we conduct an ablation study in Sec.\ref{sec:ablation}. 
% More quantitative and qualitative results can be found in the supplemental materials.

\vspace{\subsubsecVspaceUp}
\subsubsection{Quantitative Comparisons\label{sec:quantitative}}
\vspace{\subsubsecVspace}
% In Table~\ref{table:quantity}, the best results are marked in {\color{red}red}, and the second best results are in {\color{blue}blue}. The percentage in the bracket indicates the improvements over the second best results. Therefore, the percentage of the best results are negative, the second best are all zeros while the others are positive. `$I_{3\times3}$' refers to no alignment and `Gyro Field' refers to alignment with pure gyro data. 

In Table~\ref{table:quantity}, the best results are marked in red, and the second-best results are in blue. The percentage in the bracket indicates the improvements over the second-best results. Therefore, the percentage of the best results is negative. The second best is all zeros while the others are positive. `$I_{3\times3}$' refers to no alignment, and `Gyro Field' refers to alignment with pure gyro data. 

For traditional methods, we compare our GyroFlow with DIS~\cite{kroeger2016fast} and DeepFlow~\cite{weinzaepfel2013deepflow} pre-trained on Sintel~\cite{butler2012naturalistic} (Table~\ref{table:quantity}, 3$\sim$4). As seen, their average EPEs are $4$ times larger than ours. In particular, DIS fails in foggy scenarios, and DeepFlow crashes in rainy scenes. Moreover, we try to implement the traditional gyroscope-based optical flow method~\cite{li2018efficient} because no replies are received from the authors. Due to the lack of implementation details, we do not get reasonable results, so they are not reported.

Next, we compare with deep supervised optical flow methods, including FlowNet2~\cite{dosovitskiy2015flownet}, IRRPWC~\cite{hur2019iterative}, SelFlow~\cite{liu2019selflow}, and recent state-of-the-art method RAFT~\cite{teed2020raft} (Table~\ref{table:quantity}, line $5\sim8$). For the lack of ground-truth labels during training, we cannot refine these methods on our trainset. So for each method, we search different pre-trained models and test them on the evaluation set. Here, we only report the best results. RAFT pre-trained on FlyingChairs~\cite{dosovitskiy2015flownet} performs the best, but it is still not as good as ours. 

We also compare our method to deep unsupervised optical flow methods, including DDFlow~\cite{liu2019ddflow}, UnFlow~\cite{meister2018unflow}, ARFlow~\cite{liu2020learning} and UFlow~\cite{jonschkowski2020matters} (Table~\ref{table:quantity}, $9\sim12$).
Here, we refine the models on our training set. UFlow achieves 3 second-best results. However, it is still not comparable with ours due to the unstable performance in challenging scenes.

As discussed in Sec.~\ref{sec:related}, RainFlow~\cite{li2019rainflow} is designed to estimate the optical flow under rainy scenes. FogFlow~\cite{yan2020optical} aims for foggy environments, and DarkFlow~\cite{zheng2020optical} intends to compute flows in the low-light scenarios. We also compare these methods. Note that all these methods are not open source. For DarkFlow~\cite{zheng2020optical}, the authors do not provide source codes but offer a pre-trained version on FlyingChairs, the result is reported at line $13$ in Table~\ref{table:quantity}. For the other two methods~\cite{li2019rainflow,yan2020optical}, no replies are received from the authors. We try to implement them, but the results are not satisfactory due to the uncertainty of some implementation details. Therefore, results are not illustrated in Table~\ref{table:quantity}.

We find that our GyroFlow model is robust in all scenes and computes a $0.717$ EPE error which is $26.46\%$ better than the second-best method on average. Notably, for `Dark' scenes that consist of poor image texture, the `Gyro Field' alone achieves the second-best performance, indicating the importance of incorporating gyro motion, especially when the image contents are not reliable. 

\begin{table*}[h]
\small
	\centering
    \resizebox*{0.98\linewidth}{!}
	{
	\begin{tabular}{>{\arraybackslash}p{5cm}
    >{\centering\arraybackslash}p{2.4cm}
    >{\centering\arraybackslash}p{2.4cm}
    >{\centering\arraybackslash}p{2.4cm}
    >{\centering\arraybackslash}p{2.4cm}
    >{\centering\arraybackslash}p{2.4cm}}
		\toprule  
		Method &RE& Dark& Fog& Rain& Avg \\ 
		
		\midrule  
		1) $\mathcal{I}_{3 \times 3}$ & 4.962(+457.53\%)  & 3.278(+228.13\%)   & 7.358(+643.23\%)   & 5.567(+425.68\%)   & 5.665(+481.03\%)\\
		
 		\midrule  
 		2) Gyro Field & 2.583(+190.22\%)  & \textcolor[rgb]{0,0,1}{0.999(+0.00\%)}   & 1.279(+29.19\%)   & 1.703(+60.81\%)   & 1.922(+97.13\%)\\
 		
 		\midrule  
		3) DIS~\cite{kroeger2016fast} & 2.374(+166.74\%)      & 2.442(+144.44\%)   & 4.677(+372.42\%)  & 3.004(+183.66\%)    & 3.399(+248.62\%)  \\
		
		4) DeepFlow~\cite{weinzaepfel2013deepflow} - Sintel\cite{butler2012naturalistic} & 3.521(+295.62\%)      & 3.425(+242.84\%)   & 3.029(+205.96\%)  & 11.812(+1015.39\%)    & 4.858(+398.26\%)  \\
		
		\midrule  
		5) FlowNet2~\cite{dosovitskiy2015flownet} - Sintel\cite{butler2012naturalistic} & 11.140(+1151.69\%)      & 44.641(+4368.57\%)   & 2.633(+165.96\%)  & 5.767(+444.57\%)    & 6.701(+587.28\%)  \\
		
		6) IRRPWC~\cite{hur2019iterative} - FlyingChairs~\cite{dosovitskiy2015flownet} & 12.487(+1303.03\%)      & 69.864(+6893.39\%)   & 1.916(+93.54\%)  & 9.799(+825.31\%)    & 8.234(+744.51\%)  \\
		
		7) SelFlow\cite{liu2019selflow} - Sintel~\cite{butler2012naturalistic} & 4.186(+370.34\%)      & 2.747(+174.97\%)   & 7.307(+638.08\%)  & 4.787(+352.03\%)    & 5.626(+477.03\%)  \\
		
		8) RAFT~\cite{teed2020raft} - FlyingChairs\cite{dosovitskiy2015flownet} & 1.246(+40.00\%)      & 1.297(+29.83\%)   & 1.136(+14.75\%)   & 1.187(+12.09\%)    & 1.349(+38.36\%)  \\
% 		5) Raft(Kitti)~\cite{} & 1.273      & 58.764   & 2.832  & 7.995    & 6.556  \\
		
% 		8) IrrPwc(Kitti)~\cite{} & 7.721      & 58.57   & 5.647  & 22.624    & 12.564  \\
		
% 		11) DeepFlow(Kitti)~\cite{} & 12.903      & 48.003   & 4.772  & 18.628    & 10.412  \\

% 		13) EpicFlow()~\cite{} & todo     & todo   & todo  & todo    & todo  \\
		\midrule  
		
% 		14) DDFlow(Kitti)~\cite{} & 9.352     & 30.67   & 8.07  & 13.659    & 9.879  \\
		
% 		10) DDFlow~\cite{liu2019ddflow}(FlyingChairs~\cite{dosovitskiy2015flownet}) & 9.352(+950.79\%)     & 17.263(+1628.03\%)   & 5.250(+430.30\%)  & 7.347(+593.77\%)    & 5.631(+477.54\%)  \\
		
% 		11) SelFLow~\cite{liu2019selflow}(Sintel\cite{butler2012naturalistic}) & 4.186(+370.34\%)     & 2.747(+174.97\%)   & 7.307(+638.08\%)  & 4.787(+352.03\%)    & 5.626(+477.03\%)  \\

% 		12) ARFlow~\cite{liu2020learning}(Kitti 2015\cite{menze2015object}) & 1.145(+28.65\%)     & 1.739(+74.07\%)   & 1.725(+74.24\%)  & 3.775(+256.47\%)    & 2.525(+158.97\%)  \\
		
% 		16) ARFlow(Kitti 2012)~\cite{liu2020learning} & 2.165     & 3.355   & 2.802  & 6.413    & 4.568  \\
		
% 		16) ARFlow(CitySpaces)~\cite{liu2020learning} & 2.342     & 5.500   & 1.965  & 6.108    & 3.784  \\
		
% 		16) ARFlow(Sintel)~\cite{liu2020learning} & 1.579     & 8.571   & 1.439  & 5.595    & 4.634  \\
		
% 		12) UFlow~\cite{jonschkowski2020matters}(todo\cite{}) & -     & -   & -  & -    & -  \\

        9) DDFlow~\cite{liu2019ddflow} - GOF & 2.273(+155.39\%)     & 2.843(+184.58\%)   & 3.070(+210.10\%)  & 2.422(+128.71\%)    & 2.527(+159.18\%)  \\
		
		10) UnFlow~\cite{meister2018unflow} - GOF & 1.120(+25.84\%)     & 1.671(+67.17\%)   & \textcolor[rgb]{0,0,1}{0.990(+0\%)}  & 1.343(+26.53\%)    & 1.221(+25.13\%)  \\
		
		11) ARFlow~\cite{liu2020learning} - GOF & 0.972(+9.21\%)     & 1.205(+20.62\%)   & 1.186(+19.80\%)  & 1.093(+3.21\%)    & 1.035(+6.15\%)  \\
		
		12) UFlow~\cite{jonschkowski2020matters} - GOF & \textcolor[rgb]{0,0,1}{0.890(+0.00\%)}     & 1.641(+64.26\%)   & 0.994(+0.40\%)  & \textcolor[rgb]{0,0,1}{1.059(+0.00\%)}    & \textcolor[rgb]{0,0,1}{0.975(+0.00\%)}  \\
		\midrule  
		
		13) DarkFlow~\cite{zheng2020optical} - FlyingChairs\cite{dosovitskiy2015flownet} & 4.127(+363.71\%)    & 4.346(+335.04\%)   & 7.316(+638.99\%)  & 4.891(+361.85\%)    & 5.758(+490.56\%)  \\
		
% 		14) RainFlow~\cite{li2019rainflow} & - & - & - & - & - \\
		
% 		15) FogFlow~\cite{yan2020optical} & - & - & - & - & - \\
		
		\midrule  
		
		14) Ours &\textcolor[rgb]{1,0,0}{0.742(\textminus16.63\%)}    & \textcolor[rgb]{1,0,0}{0.902(\textminus9.71\%)}   &\textcolor[rgb]{1,0,0}{0.658(\textminus33.54\%)}  &\textcolor[rgb]{1,0,0}{0.730(\textminus31.07\%)}    &\textcolor[rgb]{1,0,0}{0.717(\textminus26.46\%)}  \\
		\bottomrule  
	\end{tabular}
	}
	\vspace{\tabVspaceUp}
	\caption{Quantitative comparisons on the evaluation dataset. We mark the best performance in red and the second-best in blue. The percentage in the bracket indicates the improvements over second-best results. We use '-' to indicate which dataset the model is trained on.}
\label{tab:results}
\label{table:quantity}
\vspace{\tabVspaceDown}
\end{table*}

For the further comparison to supervised methods, we expand the evaluation set to $400$ pairs, then it is divided into $2$ parts, GOF-clean (for training) and GOF-final (for testing). We pre-train the supervised methods on FlyingChairs~\cite{dosovitskiy2015flownet}, then fine-tune them on GOF-clean. We also fine-tune UFlow~\cite{jonschkowski2020matters} and GyroFlow on GOF-clean. Results evaluated on GOF-final are shown in Table~\ref{table:sup_comnpare}. As seen, for unsupervised methods, we are better than UFlow. For supervised methods, we are better than FlowNet2~\cite{dosovitskiy2015flownet} and IRRPWC~\cite{hur2019iterative}. RAFT~\cite{teed2020raft} achieves the best on average. Note that, supervised methods have label guidance during the entire training while we do not.
% Our method can further improve the quality according to the reliable parts from these unreliable scenes.  

\begin{table}[h]
    \small
    \centering
    \resizebox*{0.98\linewidth}{!}{
    \begin{tabular}{lllllllll}
        \toprule  
        Model & RE & Dark & Fog & Rain & Avg \\
        \midrule
        FlowNet2~\cite{dosovitskiy2015flownet} - GOF-clean & 0.67 & 4.74 & 5.21 & 3.73 & 3.36 \\
        IRRPWC~\cite{hur2019iterative} - GOF-clean & 0.64 & 5.00 & 5.00 & 4.40 & 3.62 \\  
        RAFT~\cite{teed2020raft} - GOF-clean  & 0.14 & 1.20 & 0.88 & 1.33 & 0.74 \\
        \midrule
        UFlow~\cite{jonschkowski2020matters} - GOF-clean & 0.72 & 3.54 & 1.50 & 3.51 & 2.37 \\
        \midrule
        Ours & 0.64 & 2.50 & 0.55 & 3.03 & 1.78\\
        \bottomrule
    \end{tabular}}
    \vspace{\tabVspaceUp}
    \caption{Comparisons on GOF-final ($200$ pairs). We use '-' to indicate which dataset the model is fine-tune on.}
    \label{table:sup_comnpare}
    \vspace{\tabVspaceDown}
\end{table}

%For the first $2$ methods, we didn't find source code nor receive the feedback by authors after sending emails. For DarkFlow, the author didn't provide source codes, but offer a pretrained version on FlyingChairs. The results show that our method is much better than DarkFlow including low light scenes. For the last part, we refine DDFlow~\cite{liu2019ddflow}, ARFlow~\cite{liu2020learning} and UFlow~\cite{jonschkowski2020matters} on our training set, note that UFLow is the state-of-art unsupervised method. 

\vspace{\subsubsecVspaceUp}
\subsubsection{Qualitative Comparisons\label{sec:qualitative}}
\vspace{\subsubsecVspace}
% In Fig.~\ref{fig:qualitative}, we illustrate the qualitative results on the evaluation set. We choose $4$ examples from $4$ different scenes inclding the low-light scene (Dark), the foggy scene (Fog), the rainy scene (Rain), and the regular scene (RE). For the methods, we choose to compare with gyro field and $2$ recent unsupervised methods, i.e., ARFlow~\cite{liu2020learning} and the state-of-art framework UFlow~\cite{jonschkowski2020matters} which are refined on our training set. In Fig.~\ref{fig:qualitative}, we show both flows and the corresponding error maps and also report the EPE errors for each example. As shown, for challenge cases, our method can fuse the background motion from gyro field and motion of dynamic objects from image-based optical flows, delivering both better visual quality and lower EPE errors. 

In Fig.~\ref{fig:qualitative}, we illustrate the qualitative results on the evaluation set. We choose one example for each of four different scenes, including the low-light scene (Dark), the foggy scene (Fog), the rainy scene (Rain), and the regular scene (RE). To compare methods, we choose the gyro field and $2$ recent unsupervised methods, i.e., ARFlow~\cite{liu2020learning} and UFlow~\cite{jonschkowski2020matters} which are refined on our training set. In Fig.~\ref{fig:qualitative}, we show optical flow along with corresponding error maps and also report the EPE error for each example. As shown, for challenge cases, our method can fuse the background motion from the gyro field with the motion of dynamic objects from the image-based optical flow, delivering both better visual quality and lower EPE errors. 

The unsupervised optical flow methods~\cite{liu2019ddflow,liu2020learning,jonschkowski2020matters} are supposed to work well in RE scenes given sufficient texture. However, we notice that, even for the RE category, our method outperforms the others, especially at the motion boundaries. With the help of the gyro field that solves the global motion, the network can focus on challenging regions. As a result, our method still achieves better visual quality and produces lower EPE errors in RE scenarios.

\begin{table}[h]
\small
\centering
\resizebox*{0.98\linewidth}{!}{
\begin{tabular}{llllll}
\toprule
Method                   & RE   & Dark & Fog  & Rain & Avg  \\
% \midrule
% Baseline & 0.79 & 1.71 & 1.35 & 1.06 & 0.95\\
\midrule
DWI      & 3.77 & 3.15 & 5.59 & 4.24 & 4.38 \\
DPGF & 0.95 & 1.67 & 1.32 & 0.89 & 0.98 \\
\midrule
SGF-Fuse            & \textcolor[rgb]{1,0,0}{0.72} & \textcolor[rgb]{0,0,1}{0.99} & 0.99 & 0.94 & \textcolor[rgb]{0,0,1}{0.80} \\ 
SGF-Map         & 1.07 & 1.02 & 1.19 & \textcolor[rgb]{1,0,0}{0.70} & 0.90 \\
SGF-Dense              & 0.77 & 1.69 & \textcolor[rgb]{0,0,1}{0.87} & 1.00 & 0.89 \\
\midrule
GyroFlow without SGF & 0.79 & 1.71 & 1.35 & 1.06 & 0.95\\
Our SGF                      & \textcolor[rgb]{0,0,1}{0.74} & \textcolor[rgb]{1,0,0}{0.90} & \textcolor[rgb]{1,0,0}{0.66} & \textcolor[rgb]{0,0,1}{0.73} & \textcolor[rgb]{1,0,0}{0.72}\\
\bottomrule
\end{tabular}}
\vspace{\tabVspaceUp}
\caption{Comparison with alternative designs of the SGF module.}
\label{table:ablation_2}
\vspace{\tabVspaceDown}
\end{table}

\subsection{Ablation Studies\label{sec:ablation}}
\vspace{\subsecVspace}
% To evaluate the effectiveness of the design with respect to each module, we conduct ablation experiments on our evaluation set. The EPE errors are reported under $4$ sub categories, including Dark, Fog, Rain and RE, along with the average score over all categories.
% \vspace{-1mm}

To evaluate the effectiveness of the design for each module, we conduct ablation experiments on the evaluation set. EPE errors are reported under $5$ categories, including Dark, Fog, Rain, and RE, along with the average error.

\vspace{\subsubsecVspaceUp}
\subsubsection{The Design of SGF} 
\vspace{\subsubsecVspace}

\begin{figure*}[h]
\begin{center}
  \includegraphics[width=0.98\linewidth]{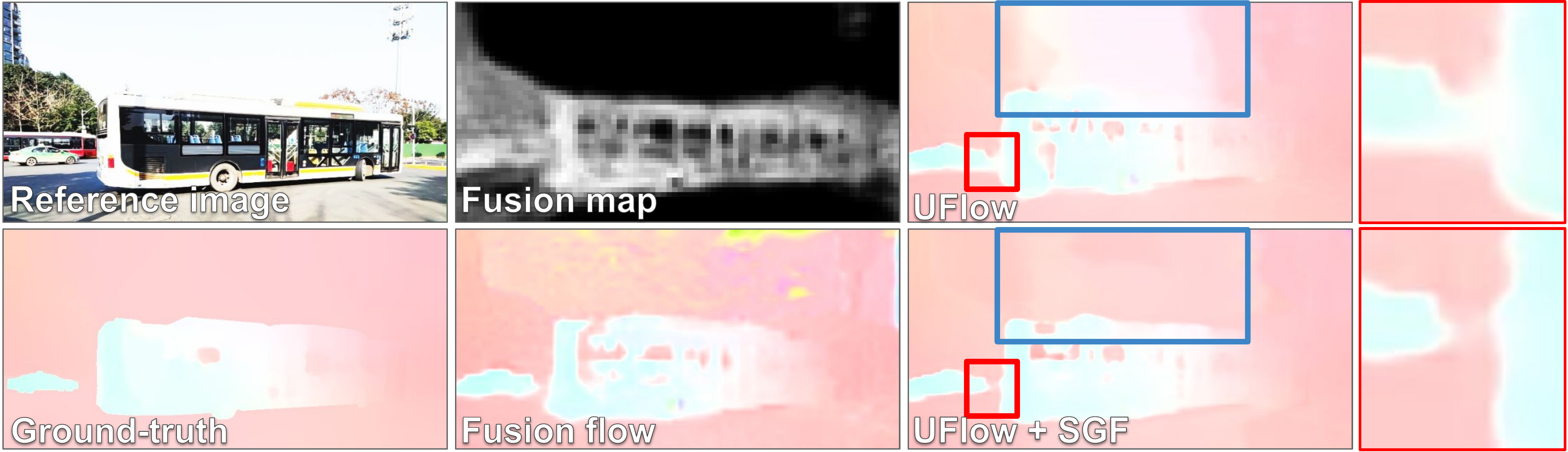}
\end{center}
\vspace{\figVspaceUp}
 \caption{Visual example of our self-guided fusion module(SGF). Results of UFlow and UFlow with SGF are shown. The fusion map is used to guide the network to focus on motion details.}
\label{fig:SGF-quatitative}
\vspace{\figVspaceDown}
\end{figure*}

For SGF, we test several designs and report results in Table~\ref{table:ablation_2}. First of all, two straightforward methods are adopted to build the module. DWI refers that we directly warp the $I_{a}$ with gyro field, then we input the warped image and $I_{b}$ to produce a residual optical flow. DPGF denotes that, for each pyramid layer, we directly add the gyro field onto the optical flow. As shown in Table~\ref{table:ablation_2}, for DWI, the result is not good. Except for the absence of gyroscope guidance during training, another possibility is that the warping operation breaks the image structure such as blurring and noising. DPGF gets a better result but is still not comparable to our SGF design because the gyro field registers background motion that should not be concatenated to dynamic object motion. Furthermore, we compare our SGF with three variants: (1) SGF-Fuse, we remove the map block, and the final fusion procedure. Although it computes a $0.8$ EPE error, it performs unstable in challenging scenes; (2) SGF-Map, where the fusion block is removed. It results in worse performance because the fusion map $M_{ab}$ tends to be inaccurate except for the rainy scene. (3) SGF-Dense, we integrate the two blocks into one unified dense block, which produces a $3$ channels tensor of which the first two channels represent the fusion flow $O_{ab}$, and the last channel denotes the fusion map $M_{ab}$. Our SGF is much better on average.

\begin{table}[]
    \small
    \centering
    \resizebox*{0.98\linewidth}{!}{
    \begin{tabular}{lllllllll}
        \toprule  
        Model & RE & Dark & Fog & Rain & Avg \\
        \midrule
        UnFlow~\cite{meister2018unflow} & 1.12 & 1.67 & 0.99 & 1.34 & 1.22 \\
        UnFlow~\cite{meister2018unflow} + SGF & 0.83 & 1.33 & 0.94 & 0.94 & 0.90 \\
        \midrule
        ARFlow~\cite{liu2020learning}  & 0.97 & 1.21 & 1.19 & 1.09 & 1.04 \\
        ARFlow~\cite{liu2020learning} + SGF & \textcolor[rgb]{0,0,1}{0.77} & 1.54 & 0.85 & 0.94 & 0.86 \\
        \midrule
        UFlow~\cite{jonschkowski2020matters} & 0.89 & 1.64 & 0.99 & 1.06 & 0.98\\
        UFlow~\cite{jonschkowski2020matters} + SGF & 0.89 & \textcolor[rgb]{0,0,1}{0.95} & \textcolor[rgb]{0,0,1}{0.71} & \textcolor[rgb]{0,0,1}{0.78} & \textcolor[rgb]{0,0,1}{0.80}\\
        \midrule
        Our baseline & 0.79 & 1.71 & 1.35 & 1.06 & 0.95\\
        Ours & \textcolor[rgb]{1,0,0}{0.74} & \textcolor[rgb]{1,0,0}{0.90} & \textcolor[rgb]{1,0,0}{0.66} & \textcolor[rgb]{1,0,0}{0.73} & \textcolor[rgb]{1,0,0}{0.72}\\
        \bottomrule
    \end{tabular}}
    \vspace{\tabVspaceUp}
    \caption{Comparison with unsupervised methods when equipped with our SGF module.}
    \label{table:ablation_1}
    \vspace{\tabVspaceDown}
\end{table}

\vspace{\subsubsecVspaceUp}
\subsubsection{Unsupervised Methods with SGF.} 
\vspace{\subsubsecVspace}
We insert the SGF module into unsupervised methods~\cite{meister2018unflow,liu2020learning,jonschkowski2020matters}, and the baseline represents our GyroFlow without SGF. In particular, similar to Fig.~\ref{fig:pipeline}, we add the SGF before the decoder $\mathbf{D}$ for each pyramid layer. Several unsupervised methods are trained on our dataset, and we report EPE errors in Table~\ref{table:ablation_1}. After inserting our SGF module into these models, noticeable improvements can be observed in Table~\ref{tab:results} and Table~\ref{table:ablation_1}, which proves the effectiveness of our proposed SGF module. Fig.~\ref{fig:SGF-quatitative} shows an example. Both background motion and boundary motion are improved after integrating our SGF.

\vspace{\subsubsecVspaceUp}
\subsubsection{Gyro Field Fusion Layer} 
\vspace{\subsubsecVspace}
Intuitively, it is also possible to fuse the gyro field only once during the training, so we add our SGF module to a specific pyramid layer. As illustrated in Table~\ref{table:ablation_3}, we notice that the more bottom layer we add SGF to, the lower EPE error it produces. The best results can only be obtained when we add the gyro field at all layers.

\begin{table}[]
\small
\centering
\resizebox*{0.98\linewidth}{!}{
\begin{tabular}{llllll}
\toprule
Pyramid Layer & RE   & Dark & Fog  & Rain & Avg  \\ 
\midrule
Baseline & \textcolor{blue}{0.79} & 1.71 & 1.35 & 1.06 & 0.95\\
% Inital Flow  & 1.02 & 1.11 & 0.86 & 0.94 & 0.94 \\ 
\midrule
$1/32$ resolution  & 1.03 & 1.04 & \textcolor[rgb]{0,0,1}{0.85} & 1.03 & 0.95 \\
$1/16$ resolution & 0.94 & \textcolor[rgb]{0,0,1}{0.98} & 0.95 & 0.93 & 0.92 \\
$1/8$ resolution  & 0.89 & 1.17 & 1.19 & \textcolor[rgb]{0,0,1}{0.87} & 0.89 \\
%Fourth Layer & 0.90 & 1.22 & 1.52 & 1.20 & 1.07 \\ 
%\midrule
$1/4$ resolution & 0.81 & 1.13 & 0.94 & 0.91 & \textcolor[rgb]{0,0,1}{0.87} \\ 
\midrule
All resolutions   & \textcolor[rgb]{1,0,0}{0.74} & \textcolor[rgb]{1,0,0}{0.90} & \textcolor[rgb]{1,0,0}{0.66} & \textcolor[rgb]{1,0,0}{0.73} & \textcolor[rgb]{1,0,0}{0.72}\\
\bottomrule
\end{tabular}}
\vspace{\tabVspaceUp}
\caption{Adding gyro filed to different pyramid layers. The baseline indicates GyroFlow without SGF.}
\label{table:ablation_3}
\vspace{\tabVspaceDown}
\end{table}

\section{Conclusion}
\vspace{\secVspace}
We have presented a novel framework GyroFlow for unsupervised optical flow learning by fusing the gyroscope data. We have proposed a self-guided fusion module to fuse the gyro field and optical flow. For the evaluation, we have proposed a dataset GOF and labeled $400$ ground-truth optical flow for quantitative metrics. The results show that our proposed method achieves state-of-art in all regular and challenging categories compared to the existing methods.

\vspace{-7pt}
\paragraph{Acknowledgement:} This research was supported in part by National Natural Science Foundation of China (NSFC) under grants No.61872067 and No.61720106004, in part by Research Programs of Science and Technology in Sichuan Province under grant No.2019YFH0016.

\newpage
% {\small
% \bibliographystyle{ieee_fullname}
% \bibliography{bib}
% }

\end{document}